\documentclass{hld2025} 
\usepackage{amsmath,amssymb}
\usepackage{mathtools}
\usepackage{graphicx}
\usepackage{hyperref}
\usepackage{booktabs}
\usepackage{xcolor}
\usepackage{enumitem}
\usepackage{caption}
\usepackage{algorithm}
\usepackage{algorithmic}
\usepackage{booktabs}
\usepackage{multirow}
\usepackage{wrapfig}
\usepackage{subcaption}

\graphicspath{{figures/}}

\hypersetup{
    colorlinks=true,
    linkcolor=blue!60!black,
    citecolor=green!50!black,
    urlcolor=blue!70!black
}

\DeclareMathOperator*{\argmin}{arg\,min}

\newcommand{\bout}{\beta_{\mathrm{out}}}

\newtheorem{proposition}[theorem]{Proposition}
\newtheorem{corollary}[theorem]{Corollary}
\newtheorem{lemma}[theorem]{Lemma}

\title[Outer-Momentum Restarting in High-Dimensional Two-Phase Optimization]{Outer-Momentum Restarting in High-Dimensional Two-Phase Optimization}

\hldauthor{%
\Name{Kristi Topollai} \Email{kt2664@nyu.edu}\\
\Name{Allan Ma} \Email{allan.ma@nyu.edu}\\
\Name{Tolga Dimlioglu} \Email{td2249@nyu.edu}\\
\Name{Sui Jiet Tay} \Email{st5494@nyu.edu}\\
\Name{Anna Choromanska} \Email{ac5455@nyu.edu}\\
\addr {New York University, New York, USA}
\vspace{-0.5em}
}

\begin{document}

\maketitle

\begin{abstract}

Communication-efficient distributed optimizers such as DiLoCo reduce synchronization costs by letting workers perform many local updates before aggregating their progress with an outer momentum optimizer. Recent theory
suggests that the outer optimizer acts on an effective spectrum induced by the inner optimization loop, and that the choice of outer momentum controls how progress
from local updates is accumulated across communication rounds. We study periodic restarting of the outer momentum as a simple complementary mechanism for controlling this outer memory. In a linearized squared-loss model where prediction-space residuals evolve under the empirical NTK, we derive a mode-wise restart contraction showing that resets exploit phase cancellation by discarding stale momentum while preserving inner-loop progress. Toy experiments verify the predicted contraction behavior, and language-model pretraining experiments show that periodic restarts widen the stable range of outer learning rates and momentum values across communication periods.
\end{abstract}



\section{Introduction}

Modern language-model (LM) training is increasingly constrained by communication in distributed setups. 
In standard synchronous data-parallel training, workers exchange gradients or optimizer states at every step, which becomes costly when accelerators such as GPUs or TPUs are distributed across machines or clusters. This led to renewed interest in local and 
\emph{two-phase} training methods, where workers perform several local 
optimization steps before synchronizing 
\cite{mcmahan2017communication,malinovskiy2020local,reddi2021adaptive,
charles2021convergence,charles2022iterated,khaled2025outer}. In this view, 
training is split into an inner phase, which performs independent local 
progress, and an outer phase, which aggregates that progress across workers. 
Similar to Lookahead methods \cite{zhang2019lookahead}, DiLoCo brings this structure to large-scale LM training: workers run many inner 
steps, form a pseudo-gradient from their parameter displacement, and pass it to 
an outer optimizer \cite{douillard2023diloco, douillard2025streaming}. Empirically, DiLoCo and its 
variants can substantially reduce communication while maintaining training 
quality, and recent work suggests that some benefits arise even in the 
one-worker setting \cite{douillard2023diloco,kallusky2025snoo}. This way, the 
outer optimizer determines how accumulated local progress is filtered, or damped at communication time.

Recent works on local SGD and federated optimization study how outer learning rates and momentum trade off optimization speed against stochastic noise \cite{reddi2021adaptive,charles2022iterated,khaled2025outer}. The closest theoretical lens is provided in \citet{agarwala2026twophase}, where it is
shown that the inner loop induces an effective spectrum on which the outer optimizer operates. Their analysis shows that heavy-ball outer momentum can constrain directions on which the inner loop already makes strong progress, while Nesterov outer momentum mitigates
this by making contraction depend on the effective progress of each mode
\cite{agarwala2026twophase}. This is consistent with recent empirical gains
from Nesterov pseudo-gradient updates \cite{kallusky2025snoo}.

We study momentum restarting as a complementary mechanism for two-phase optimizers. Rest-arting has a long history in accelerated optimization, where resetting momentum can reduce oscillation or correct misalignment
\cite{odonoghue2015adaptive,giselsson2014monotonicity,
fercoq2019adaptive,kim2018adaptive}, with related restart ideas being used in deep learning to improve stability \cite{wang2022scheduled,huangspam,topollai2026understanding}. In two-phase training, the outer momentum buffer stores pseudo-gradient history only at communication rounds, which means that outdated outer memory can affect how local progress is aggregated across synchronizations. We show that periodically resetting this buffer reduces fragility to the outer momentum and learning rate. Our theory explains the effect through a phase-dependent contraction, while our
\(150\)M-parameter LLaMA pretraining experiments show that restarts widen the stable region of outer-optimizer hyperparameters, a useful property in
pretraining where optimizer tuning is already delicate
\cite{wen2025fantasticpretrainingoptimizers,charles2025scaling} and two-phase methods introduce an additional outer-loop hyperparameter layer.

\section{Preliminaries: deterministic two-phase mode dynamics}
\label{sec:prelim}

We introduce a deterministic mode-wise model to isolate how the outer optimizer
processes the progress accumulated by the inner loop. To do so, we follow the two-phase analysis of \cite{agarwala2026twophase}. The object being analyzed is the training residual in prediction space,
\( R(\theta)=f(\theta)-y_{\mathrm{tr}} \)
where $f(\theta)$ denotes the vector of model predictions on the training set and
$y_{\mathrm{tr}}$ denotes the training targets. Under the standard linearized squared-loss/NTK approximation, the residual
dynamics are governed by the empirical kernel \(H\). After diagonalizing this
kernel, the residual decomposes into independent scalar modes, so we use a
deterministic scalar-mode surrogate for the residual dynamics rather than
modeling the full nonlinear parameter trajectory \cite{jacot2018neural,lee2019wide,arora2019exact}.

We analyze one residual eigencoordinate at a time. Let $\lambda\ge 0$ denote the
corresponding empirical-kernel eigenvalue. Within one communication round, the
inner optimizer performs $S$ gradient-descent steps with step size $\eta$ where $0\le \eta\lambda \le 1$. If
$x_t$ is the value of this residual mode at the beginning of outer round $t$,
then after the inner steps,
\vspace{-1mm}
\begin{equation}
x_t^{\mathrm{loc}}=(1-\eta\lambda)^Sx_t,\qquad
g_t:=x_t-x_t^{\mathrm{loc}}=\sigma x_t,\qquad
\sigma:=1-(1-\eta\lambda)^S .
\label{eq:inner_pseudograd}
\vspace{-1mm}
\end{equation}

Here $x_t^{\mathrm{loc}}$ denotes the value of the same residual mode after the worker has completed \(S\) local inner steps and
$g_t$ is the pseudo-gradient passed to the outer optimizer. 
The coefficient $\sigma\in[0,1]$ is the effective progress seen by the outer optimizer at
communication time and large $\sigma$ means that the inner loop has already made
substantial progress on that residual mode.

For heavy-ball/EMA outer momentum, let $m_t$ denote the outer momentum buffer. At outer round $t$, the pseudo-gradient $g_t$ is accumulated through $m_{t+1}=\bout m_t+(1-\bout)g_t$, and the residual eigencoordinate is updated as $x_{t+1}=x_t-\nu m_{t+1}$. Since the inner loop gives $g_t=\sigma x_t$, the pair consisting of the residual mode and its outer momentum, $z_t=(x_t,m_t)^\top$, evolves linearly across communication rounds. Substituting $g_t=\sigma x_t$ into the two scalar updates yields
\begin{equation}
z_{t+1}=T_{\mathrm{HB}}(\sigma)z_t \qquad \text{where} \qquad
T_{\mathrm{HB}}(\sigma)=
\begin{pmatrix}
1-\nu(1-\bout)\sigma & -\nu\bout\\
(1-\bout)\sigma & \bout
\end{pmatrix}.
\label{eq:transition_hb}
\vspace{-1mm}
\end{equation}
Since \(\det(T_{\mathrm{HB}})=\bout\), complex-regime eigenvalues have magnitude
\(\sqrt{\bout}\), giving the \(\sigma\)-indepen-dent damping rate
\(r_\infty=-\frac12\log\bout\). This is the outer-momentum damping effect, that is, even
high-progress modes with \(\sigma\approx1\) can be limited by the same slow HB
envelope.

For Nesterov-style outer momentum, the buffer update is unchanged, but the
outer step uses
\(x_{t+1}=x_t-\nu((1+\bout)m_{t+1}-\bout m_t)\), which gives
\begin{equation}
T_{\mathrm{NAG}}(\sigma)=
\begin{pmatrix}
1-\nu(1-\bout^2)\sigma & -\nu\bout^2\\
(1-\bout)\sigma & \bout
\end{pmatrix}, \qquad
\det(T_{\mathrm{NAG}})=\bout\bigl(1-(1-\bout)\nu\sigma\bigr).
\label{eq:transition_nag}
\end{equation}

\section{Restarting Outer Momentum: An Old Trick Revisited}
\label{sec:method}
The preceding discussion explains why Nesterov outer momentum is effective in
two-phase training. For HB, complex modes share the same envelope
\(\rho=\sqrt{\bout}\), so the outer update may remain slow even when the inner
loop has made substantial progress. NAG addresses this spectrally as its
contraction depends on \(\nu\sigma\), making the outer optimizer more responsive
to high-progress modes. We study a complementary mechanism: periodic restarting
of the outer momentum buffer. Momentum restarting is classical in accelerated
optimization, where resets can reduce oscillations or correct momentum
misalignment
\cite{odonoghue2015adaptive,giselsson2014monotonicity,
fercoq2019adaptive,kim2018adaptive,roulet2017sharpness}. In our setting, after
every \(K\) outer rounds, we reset only the outer buffer, \(m\leftarrow0\).
Unlike NAG, restarting does not modify the one-step spectrum; it modifies the
duration over which outer memory is retained. In the complex regime, the
residual coordinate oscillates within an envelope, so an appropriately
chosen restart period can stop the trajectory near a phase cancellation, where
the projected residual is much smaller than the envelope. Thus NAG and
restarting act through distinct mechanisms: NAG changes the contraction factor,
whereas restarting exploits favorable reset times. Starting a restart cycle with zero momentum, \(m\leftarrow0\), one cycle of \(K\)
outer rounds maps the initial residual mode to
\[
x_K=\chi_K(\sigma)x_0,
\qquad
\chi_K(\sigma):=\bigl[T_{\mathrm{HB}}(\sigma)^K\bigr]_{11},
\qquad
r_K(\sigma):=-\frac1K\log|\chi_K(\sigma)|.
\]
Thus \(\chi_K\) is simply the scalar contraction of the residual mode after one
restart cycle, and \(r_K\) is its average per-round contraction rate.

\begin{figure*}[h]
\centering
\includegraphics[width=\textwidth]{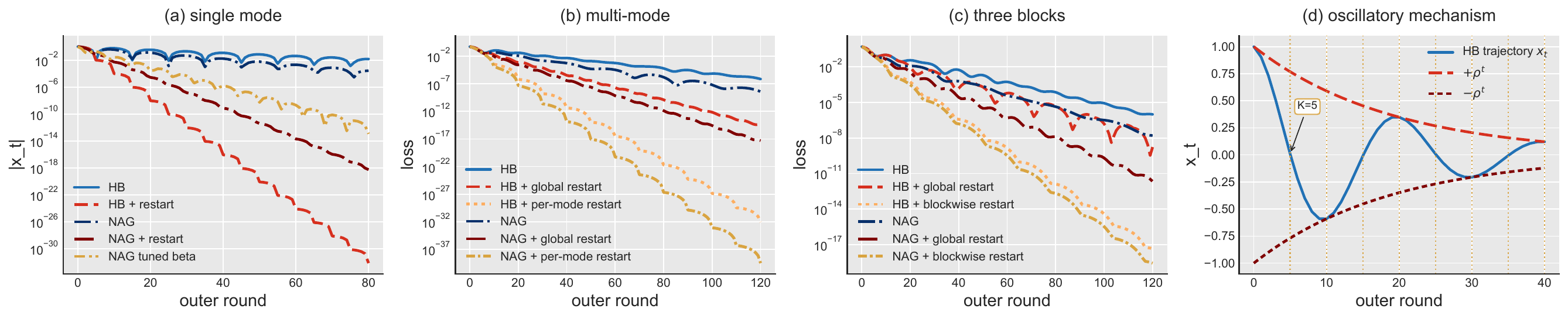}
\caption{\textbf{Restart mechanism across scalar, multi-mode, and blockwise settings.}
Periodic restart exploits phase cancellation in the outer momentum dynamics.
In the scalar case, HB with restart can match a tuned no-restart NAG baseline;
in heterogeneous spectra, global restart improves over no restart, while
per-mode or blockwise periods capture additional gains. Panel (d) visualizes
the underlying oscillatory mechanism: the residual coordinate moves inside the
HB envelope \(\pm\rho^t\), and restarting near a cancellation time makes the
projected residual small.}
\label{fig:restart_mechanism}
\vspace{-5mm}
\end{figure*}

\begin{proposition}[Restart contraction]
\label{prop:chiK}
Let \(a=1-\nu(1-\bout)\sigma\). Then
\vspace{-1.5mm}
\[
\chi_K=(a+\bout)\chi_{K-1}-\bout\chi_{K-2},
\qquad
\chi_0=1,\quad \chi_1=a.
\vspace{-1.5mm}
\]

In the complex regime, write the eigenvalues of \(T_{\mathrm{HB}}\) as
\(\rho e^{\pm i\varphi}\), where $
\rho=\sqrt{\bout},
\
\cos\varphi
=\frac{a+\bout}{2\rho}.$
Then
\vspace{-2.5mm}
\[
\chi_K(\sigma)
=
\rho^K
\Bigl[
\cos(K\varphi)+C\sin(K\varphi)
\Bigr],
\qquad
C=\frac{a-\bout}{2\rho\sin\varphi}.
\vspace{-2.5mm}
\]
\end{proposition}

For the high-momentum outer optimizers considered here, the complex regime is
not a pathological case but the typical operating regime. The condition
\(\operatorname{tr}(T_{\mathrm{HB}})^2<4\det(T_{\mathrm{HB}})\) is equivalent to
\vspace{-1mm}
\[
\frac{1-\sqrt{\bout}}{\nu(1+\sqrt{\bout})}
<
\sigma
<
\frac{1+\sqrt{\bout}}{\nu(1-\sqrt{\bout})}.
\vspace{-1mm}
\]
When \(\bout\) is close to \(1\), this interval covers essentially all
nontrivial effective progress values \(\sigma\in(0,1]\) for \(\nu=O(1)\): the
lower endpoint is close to zero, while the upper endpoint is far larger than
one. Thus, the oscillatory regime captured by Proposition~\ref{prop:chiK} is the
relevant one for typical outer momentum choices; the derivation is given in
Appendix~\ref{app:complex-regime}. In this regime, the closed form decomposes
the restarted dynamics into the usual HB envelope \(\rho^K\) and a
phase-dependent projection term. Restart improves over the non-restarted HB
envelope precisely when this projection term has magnitude below one.
\begin{proposition}[\(K\)-period crossover]
\label{prop:K_crossover}
In the complex regime, a \(K\)-period restart improves over the non-restarted
HB envelope iff
\vspace{-2mm}
\[
r_K(\sigma)>r_\infty
\quad\Longleftrightarrow\quad
|\chi_K(\sigma)|<\rho^K
\quad\Longleftrightarrow\quad
|\cos(K\varphi)+C\sin(K\varphi)|<1 .
\]
Equivalently, writing $
\cos(K\varphi)+C\sin(K\varphi)
=
\sqrt{1+C^2}\cos(K\varphi-\theta),
\
\theta=\arctan C,$
useful periods are those whose phase lands near a cancellation point,
$
K_\ell
\approx
\left\lfloor
\frac{\theta+\pi/2+\ell\pi}{\varphi}
\right\rceil,
\ \ell\in \mathbb{N}^+.$
\end{proposition}

\vspace{-5mm}
\begin{figure*}[htp]
\centering
\includegraphics[width=0.95\textwidth]{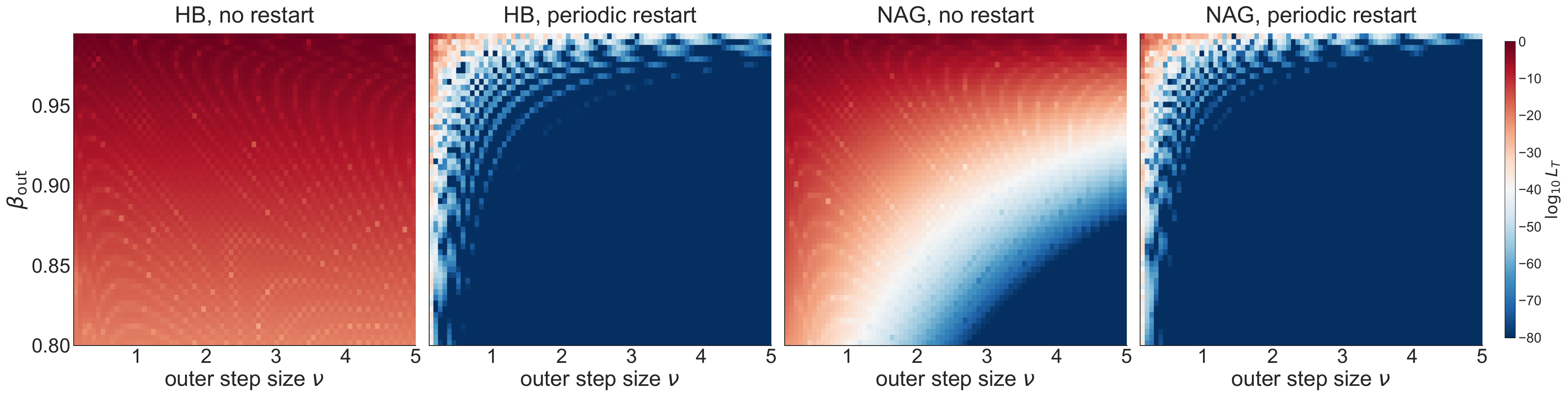}
\vspace{-2mm}
\caption{\textbf{Robustness over outer hyperparameters.}
Each heatmap shows clipped \(\log_{10}\) final loss over a grid of \(\bout\)
and \(\nu\). The best periodic restart enlarges the good-hyperparameter region for HB and NAG.}
\label{fig:robustness_toy}
\vspace{-2mm}
\end{figure*}

In experiments, we select \(K\) from an admissible integer range by minimizing
\(|\chi_K(\sigma)|\), equivalently maximizing the restarted rate \(r_K\). The
phase interpretation clarifies why no single period is uniformly optimal as large-\(\sigma\) components rotate faster and favor shorter periods, whereas small-\(\sigma\) components rotate more slowly and favor longer periods. This tradeoff is illustrated in Figure~\ref{fig:restart_mechanism}, which compares
global, per-mode, and blockwise restart periods as spectral heterogeneity increases. We apply the same projected-factor construction to NAG, and defer
the NAG recurrence, closed form, and blockwise extension to the Appendix. Figure~\ref{fig:robustness_toy} further shows that these restart periods expand the
region of outer learning rates and momentum values that yield stable
performance.

\section{ Experiments}
\label{sec:experiments}

We now present our empirical results for LM pretraining. We follow the general DiLoCo training and pretrain a 150M-parameter LLaMA model in a 2-replica DiLoCo configuration using two H200 GPUs. All runs use a fixed training budget of approximately 3.3B tokens, following the Chinchilla scaling rule for a 150M-parameter model~\citep{hoffmann2022training}. Additional details, results, ablations, and the full set of explored hyperparameter configurations are reported in Section~\ref{app:sec:experiments} of Appendix.

\paragraph{Restarts widen the stable outer-hyperparameter region.} 

\begin{figure}[htp!]
    \centering
    \begin{minipage}{0.49\textwidth}
        \centering
        \includegraphics[width=\linewidth]{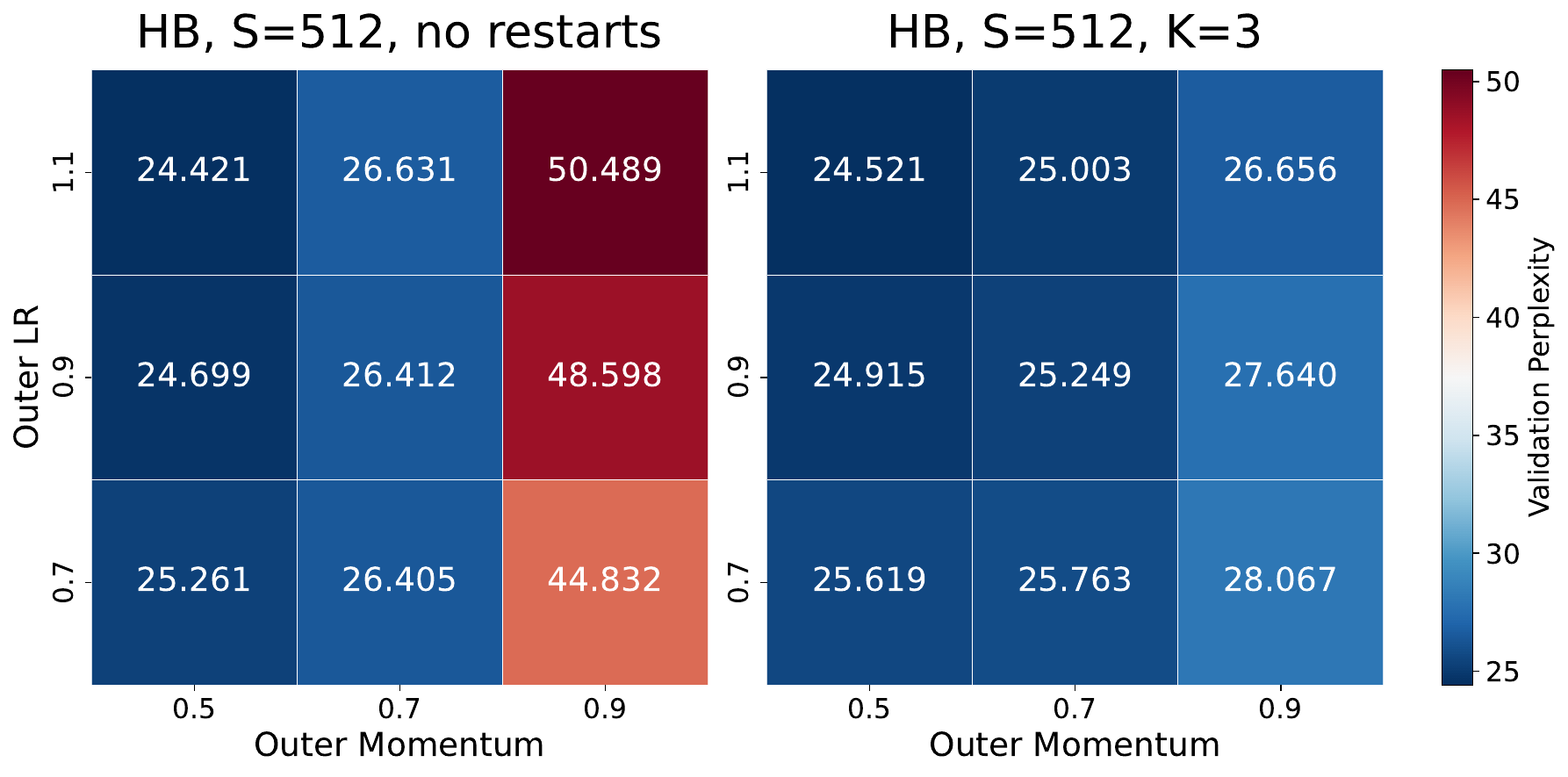}
    \end{minipage}
    \hfill
    \begin{minipage}{0.49\textwidth}
        \centering
        \includegraphics[width=\linewidth]{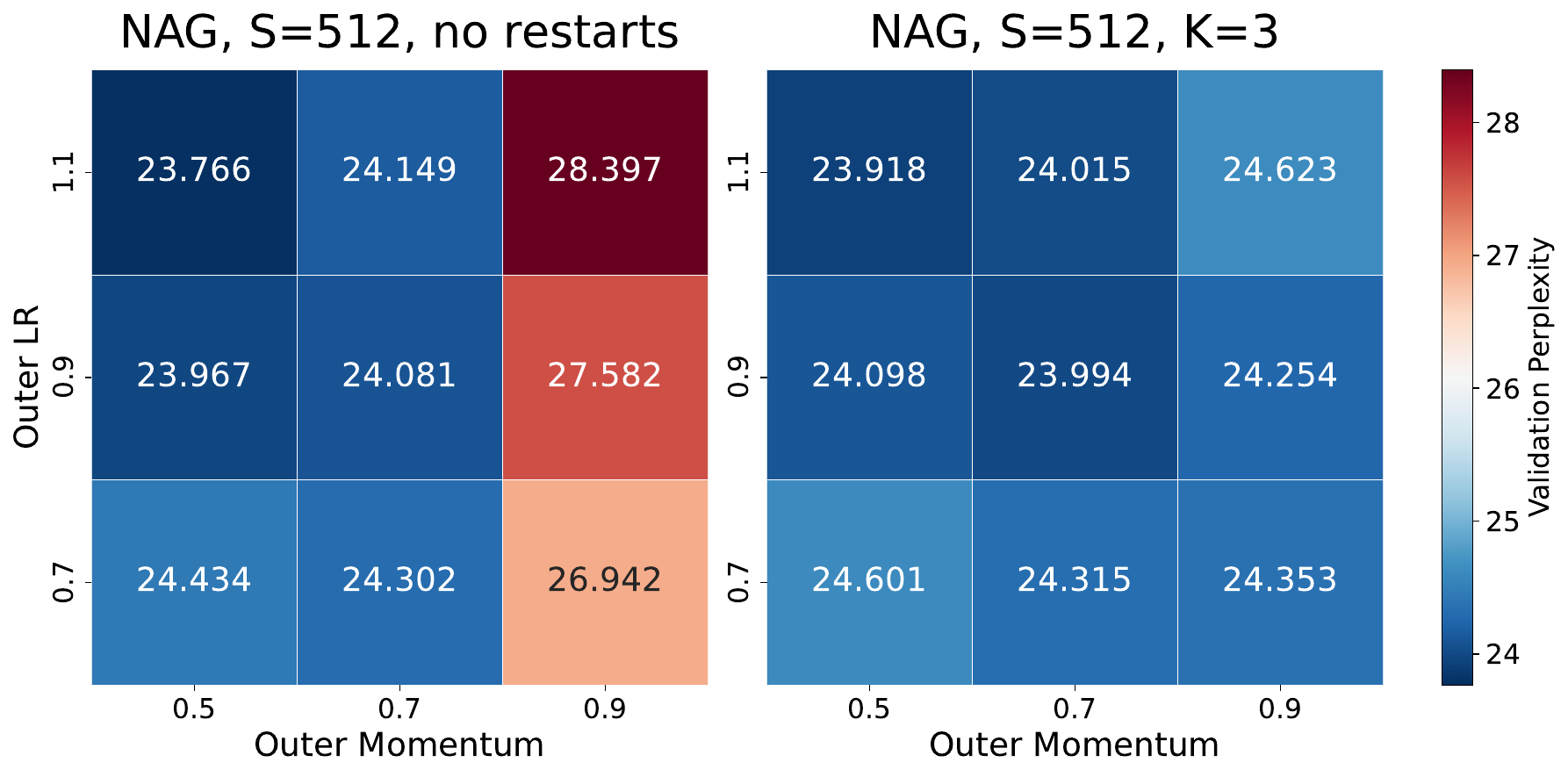}
    \end{minipage}
    \caption{Validation perplexity over the outer hyperparameter grid at \(S=512\) for HB(left) and NAG(right), comparing standard DiLoCo against DiLoCo with momentum restart period \(K=3\). Periodic restarts reduce the high-\(\beta_{\mathrm{out}}\) failure region while preserving peak performance.}
    \vspace{-1em}
    \label{fig:sens_heatmap}
\end{figure}

Figure~\ref{fig:sens_heatmap} shows validation perplexity across the outer
learning rate \(\nu\) and momentum \(\beta_{\mathrm{out}}\). Without restarts,
large \(\beta_{\mathrm{out}}\) creates a clear failure region, especially at
longer communication periods. Periodic restarts largely remove this instability
and expand the range of hyperparameters that achieve good performance. Thus the
main benefit is robustness, since restarts make the outer optimizer less sensitive to
\(\nu\) and \(\beta_{\mathrm{out}}\).

\paragraph{Restarts reduce retuning sensitivity across communication periods.}
We next test how much retuning is needed as the communication period changes.
We tune DiLoCo once at \(S=128\), \begin{wrapfigure}{r}{0.65\textwidth}
    \vspace{-3mm}
    \centering
    \begin{minipage}{0.32\textwidth}
        \centering
        \includegraphics[width=\linewidth]{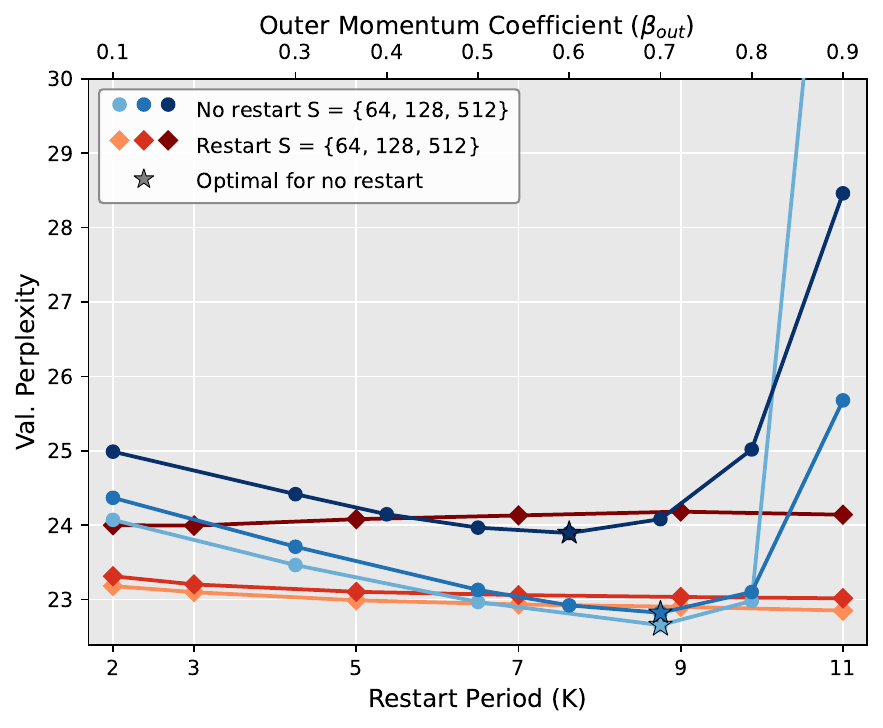}
    \end{minipage}
    \hfill
    \begin{minipage}{0.32\textwidth}
        \centering
        \includegraphics[width=\linewidth]{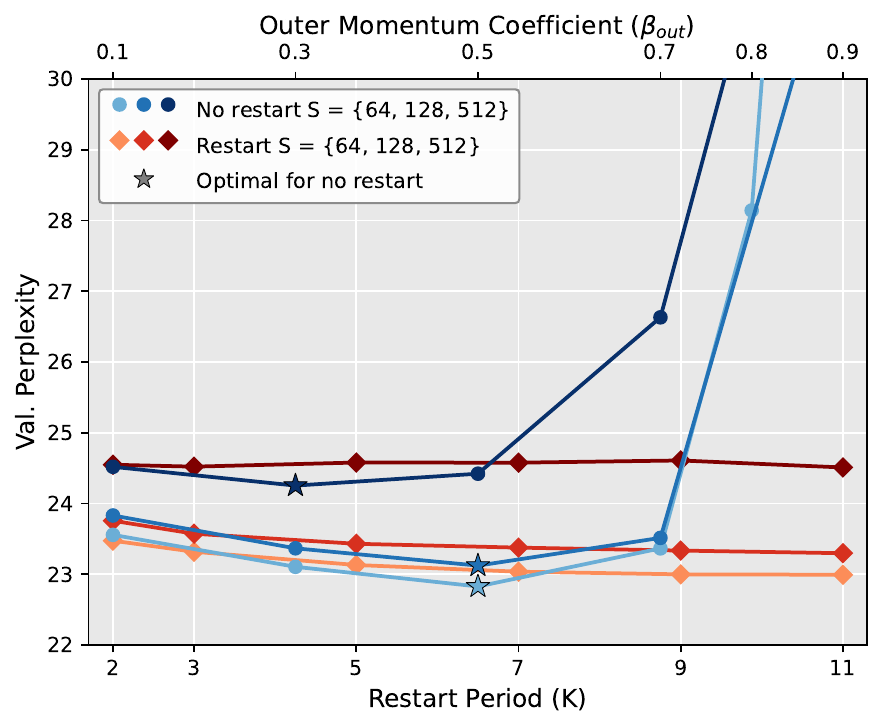}
    \end{minipage}
    \vspace{-1em}
    \caption{Final validation perplexity across communication periods
    \(S\in\{64,128,512\}\), for NAG (left) and HB (right). No-restart curves
    fix \(\nu=\nu^*\) and sweep \(\beta_{\mathrm{out}}\); restart curves fix
    \((\nu,\beta_{\mathrm{out}})=(\nu^*,\beta_{\mathrm{out}}^*)\) and sweep
    \(K\).}
    \label{fig:robust_diloco}
    \vspace{-3mm}
\end{wrapfigure} obtaining
\((\nu^*,\beta_{\mathrm{out}}^*)=(0.9,0.7)\) for NAG and
\((1.1,0.5)\) for HB. For no-restart runs, we keep \(\nu=\nu^*\) and sweep
\(\beta_{\mathrm{out}}\), while for runs with restart, we keep
\((\nu,\beta_{\mathrm{out}})=(\nu^*,\beta_{\mathrm{out}}^*)\) and sweep only
the restart period \(K\). Thus, we compare tuning the outer momentum coefficient, against tuning the
restart period. Figure~\ref{fig:robust_diloco} shows that not restarting exhibits sharp degradation at large momentum coefficients. By contrast, restart curves remain comparatively flat across \(K\), and similar periods transfer across communication lengths. Hence \(K\) serves as a more forgiving hyperparameter
than \(\beta_{\mathrm{out}}\). This effect is most pronounced for HB, where large \(\beta_{\mathrm{out}}\) can diverge, while restarts remain stable.

\section{Conclusion}
We studied periodic outer-momentum restarts for two-phase optimization and found that they can widen the stable hyperparameter region by limiting unhelpful outer memory. Although our analysis focuses on fixed-period resets, the empirical results suggest that the method is fairly insensitive to the precise period, making periodic restarting a simple and robust baseline. This opens a broader question: can the outer loop decide for itself when its memory should be refreshed? Future work should develop adaptive restart rules, per-layer or blockwise periods, and larger-scale studies of how restart schedules vary with communication period and model size.

\bibliography{restart_refs}

\appendix

\newpage

\section{Derivations for the two-phase restart dynamics}
\label{app:derivations}

\subsection{From the quadratic model to the scalar pseudo-gradient}
\label{app:inner-loop}

Consider the quadratic residual model
\[
    \ell(x)=\frac12 x^\top Hx,\qquad H\succeq0.
\]
Diagonalize \(H=U\Lambda U^\top\). Since the dynamics are linear, each
eigendirection evolves independently. For an eigenvalue \(\lambda\), \(S\)
inner GD steps with stepsize \(\eta\) give
\[
    x_t^{\mathrm{loc}}=(1-\eta\lambda)^Sx_t.
\]
The pseudo-gradient passed to the outer optimizer is the displacement produced
by the inner loop,
\[
    g_t:=x_t-x_t^{\mathrm{loc}}
    =
    \left[1-(1-\eta\lambda)^S\right]x_t
    =
    \sigma x_t,
    \qquad
    \sigma:=1-(1-\eta\lambda)^S .
\]
Thus \(\sigma\) is the effective eigenvalue observed at communication time.

\subsection{Heavy-ball/EMA outer transition}
\label{app:hb-transition}

Throughout this section, \(t\) indexes outer communication rounds. One outer
transition already includes the \(S\) inner steps, whose effect is absorbed into
\(\sigma = 1-(1-\eta\lambda)^S\).

The heavy-ball/EMA outer update is
\[
    m_{t+1}=\bout m_t+(1-\bout)g_t,\qquad
    x_{t+1}=x_t-\nu m_{t+1}.
\]
Substituting \(g_t=\sigma x_t\) gives
\[
    m_{t+1}=(1-\bout)\sigma x_t+\bout m_t,
    \qquad
    x_{t+1}
    =
    \bigl(1-\nu(1-\bout)\sigma\bigr)x_t-\nu\bout m_t.
\]
Hence, with \(z_t=(x_t,m_t)^\top\),
\[
    z_{t+1}
    =
    T_{\mathrm{HB}}(\sigma)z_t,\qquad
    T_{\mathrm{HB}}(\sigma)
    =
    \begin{pmatrix}
    1-\nu(1-\bout)\sigma & -\nu\bout\\
    (1-\bout)\sigma & \bout
    \end{pmatrix}.
\]
Let \(a:=1-\nu(1-\bout)\sigma\). Then
\[
    \operatorname{tr}(T_{\mathrm{HB}})=a+\bout,
\]
and a direct computation gives
\[
    \det(T_{\mathrm{HB}})
    =
    a\bout-(-\nu\bout)\cdot(1-\bout)\sigma
    =
    \bigl(1-\nu(1-\bout)\sigma\bigr)\bout+\nu\bout(1-\bout)\sigma
    =
    \bout.
\]
The cancellation of the \(\nu(1-\bout)\sigma\) terms is the algebraic origin of
the \(\sigma\)-independent damping rate: the determinant of the outer
transition is locked at \(\bout\), regardless of how much progress \(\sigma\)
the inner loop made.

Whenever \((a+\bout)^2<4\bout\), the two eigenvalues are complex conjugates.
Since their product is \(\bout\), they can be written as
\[
    \lambda_{\pm}
    =
    \rho e^{\pm i\varphi},
    \qquad
    \rho=\sqrt{\bout},
    \qquad
    \cos\varphi
    =
    \frac{\operatorname{tr}(T_{\mathrm{HB}})}{2\rho}
    =
    \frac{a+\bout}{2\rho}.
\]
The non-restarted asymptotic envelope is therefore \(\rho^t\), giving the
per-outer-round damping rate \(r_\infty=-\log\rho=-\frac12\log\bout\).

\subsection{Nesterov-style outer transition}
\label{app:nag-transition}

For NAG, the same momentum buffer is used, but the outer step applies the
lookahead combination
\[
    m_{t+1}=\bout m_t+(1-\bout)g_t,\qquad
    x_{t+1}=x_t-\nu\bigl((1+\bout)m_{t+1}-\bout m_t\bigr).
\]
Using \(g_t=\sigma x_t\),
\[
    (1+\bout)m_{t+1}-\bout m_t
    =
    (1-\bout^2)\sigma x_t+\bout^2m_t.
\]
Therefore
\[
    z_{t+1}
    =
    T_{\mathrm{NAG}}(\sigma)z_t,\qquad
    T_{\mathrm{NAG}}(\sigma)
    =
    \begin{pmatrix}
    1-\nu(1-\bout^2)\sigma & -\nu\bout^2\\
    (1-\bout)\sigma & \bout
    \end{pmatrix}.
\]
Its determinant is
\[
\begin{aligned}
    \det(T_{\mathrm{NAG}})
    &=
    \bout\bigl(1-\nu(1-\bout^2)\sigma\bigr)
    +\nu\bout^2(1-\bout)\sigma  \\
    &=
    \bout-\nu\bout(1-\bout)\sigma
    =
    \bout\bigl(1-(1-\bout)\nu\sigma\bigr).
\end{aligned}
\]
This dependence on \(\nu\sigma\) is the spectral advantage of NAG in the
two-phase dynamics.

\subsection{Restart contraction factor and recurrence}
\label{app:restart-recurrence}

Suppose the outer momentum is reset every \(K\) outer rounds. At the beginning
of a restart block, \(m_0=0\), so
\[
    z_0=(x_0,0)^\top=x_0e_1,\qquad e_1=(1,0)^\top.
\]
After \(K\) outer rounds of HB/EMA dynamics,
\[
    x_K
    =
    e_1^\top T_{\mathrm{HB}}(\sigma)^Kz_0
    =
    \chi_K(\sigma)x_0,
    \qquad
    \chi_K(\sigma):=e_1^\top T_{\mathrm{HB}}(\sigma)^Ke_1.
\]
The per-outer-round restarted rate is
\[
    r_K(\sigma)=-K^{-1}\log|\chi_K(\sigma)|.
\]
The characteristic polynomial of \(T_{\mathrm{HB}}\) is
\(q(\lambda)=\lambda^2-(a+\bout)\lambda+\bout\). By Cayley--Hamilton,
\[
    T_{\mathrm{HB}}^2=(a+\bout)T_{\mathrm{HB}}-\bout I.
\]
Multiplying by \(T_{\mathrm{HB}}^{K-2}\) and projecting with \(e_1^\top\) and
\(e_1\) gives
\[
    \chi_K=(a+\bout)\chi_{K-1}-\bout\chi_{K-2},
    \qquad
    \chi_0=1,\qquad
    \chi_1=a.
\]
This recurrence holds in all regimes, with real or complex eigenvalues; only
its closed-form solution depends on the regime.

\subsection{Closed form in the complex regime}
\label{app:closed-form}

In the complex regime, the eigenvalues are \(\rho e^{\pm i\varphi}\), so any
real sequence satisfying the recurrence has the form
\[
    \chi_K
    =
    \rho^K\bigl(A\cos(K\varphi)+B\sin(K\varphi)\bigr).
\]
The initial condition \(\chi_0=1\) gives \(A=1\). The condition \(\chi_1=a\)
then gives
\[
    a
    =
    \rho\bigl(\cos\varphi+B\sin\varphi\bigr),
    \qquad
    B
    =
    \frac{a/\rho-\cos\varphi}{\sin\varphi}
    =
    \frac{a-\bout}{2\rho\sin\varphi},
\]
where the second equality uses \(\rho\cos\varphi=(a+\bout)/2\). Thus
\[
    \boxed{
    \chi_K(\sigma)
    =
    \rho^K\left[
    \cos(K\varphi)+
    C\sin(K\varphi)
    \right]},
    \qquad
    C=\frac{a-\bout}{2\rho\sin\varphi}.
\]
The factor \(\rho^K\) is the same envelope as the non-restarted method, while
the bracket is a phase-dependent projection onto the iterate coordinate.

\subsection{When are we in the complex regime?}
\label{app:complex-regime}

The closed form above and most of our analysis assumes the complex regime,
\((a+\bout)^2<4\bout\). Substituting \(a=1-\nu(1-\bout)\sigma\), this becomes
\[
    \bigl(1+\bout-\nu(1-\bout)\sigma\bigr)^2<4\bout
    \quad\Longleftrightarrow\quad
    \left|1+\bout-\nu(1-\bout)\sigma\right|<2\sqrt{\bout}.
\]
The right-hand inequality factors as
\(\nu(1-\bout)\sigma<1+\bout+2\sqrt{\bout}=(1+\sqrt{\bout})^2\); the left-hand
one as \(\nu(1-\bout)\sigma>(1-\sqrt{\bout})^2\). Using
\(1-\bout=(1-\sqrt{\bout})(1+\sqrt{\bout})\), this simplifies to
\[
    \frac{1-\sqrt{\bout}}{\nu(1+\sqrt{\bout})}
    <
    \sigma
    <
    \frac{1+\sqrt{\bout}}{\nu(1-\sqrt{\bout})}.
\]
For \(\bout\) close to \(1\), the lower bound is small (of order
\((1-\bout)/(4\nu)\)) and the upper bound is large (of order
\(4/(\nu(1-\bout))\)). For example, at \(\bout=0.9\) and \(\nu=1\) the regime
is \(\sigma\in(0.026,38)\); at \(\bout=0.99\) and \(\nu=1\) it is
\(\sigma\in(0.0025,398)\). Effectively all practically relevant \(\sigma\in(0,1]\)
fall in the complex regime for the outer momenta typically used in DiLoCo.

In the boundary case \((a+\bout)^2=4\bout\) (critical damping), the eigenvalues
collide at \(\rho\); outside the complex regime, they are real and distinct,
so the iterate decays without oscillation and no \(K\) zeros the bracket. Outside the complex regime, the eigenvalues are real and distinct, so the
iterate no longer exhibits the oscillatory phase-cancellation mechanism analyzed
above. Thus, the restart benefit predicted by this phase argument is not expected
in the same way in the real-eigenvalue regime, consistent with the standard
intuition that restarting is most useful when momentum induces oscillatory
transients.

\subsection{Oracle restart period}
\label{app:oracle-period}

The single-mode oracle period minimizes the restart contraction factor:
\[
    K^\star(\sigma)
    \in
    \argmin_{K\in\mathbb{N}}|\chi_K(\sigma)|.
\]
Since \(\rho^K\) varies smoothly with \(K\), the main cancellation mechanism is
the oscillatory bracket. Write
\[
    \cos(K\varphi)+C\sin(K\varphi)
    =
    \sqrt{1+C^2}\cos(K\varphi-\theta),
    \qquad
    \theta=\arctan C.
\]
A zero of the bracket occurs when
\[
    K\varphi-\theta=\frac{\pi}{2}+\ell\pi,\qquad \ell\in\mathbb{Z}.
\]
The first positive near-cancellation is therefore approximated by
\[
    K^\star(\sigma)
    \approx
    \left\lfloor
    \frac{\theta+\pi/2}{\varphi}
    \right\rceil,
\]
where \(\lfloor\cdot\rceil\) denotes rounding to the nearest positive integer.
The rounding error costs at most a factor of
\(\sin(\varphi/2)\sqrt{1+C^2}\approx(\varphi/2)\sqrt{1+C^2}\) in the bracket
amplitude, which is small for \(\bout\) close to \(1\) (since then
\(\varphi\) is small). The phase
interpretation also explains the qualitative dependence on
\(\sigma\): increasing \(\sigma\) increases the rotation frequency
\(\varphi\), so high-\(\sigma\) modes prefer shorter restart periods.

\subsection{NAG restart factor and closed form}
\label{app:nag-restart-factor}

For NAG, we use the same projected-factor definition:
\[
    \chi_K^{\mathrm{NAG}}(\sigma)
    :=
    e_1^\top T_{\mathrm{NAG}}(\sigma)^Ke_1,
    \qquad
    r_K^{\mathrm{NAG}}(\sigma)
    :=
    -\frac1K\log|\chi_K^{\mathrm{NAG}}(\sigma)|.
\]
Let \(a_{\mathrm{N}}:=1-\nu(1-\bout^2)\sigma\) and
\(D_{\mathrm{N}}:=\det(T_{\mathrm{NAG}})=\bout(1-(1-\bout)\nu\sigma)\). The
characteristic polynomial of \(T_{\mathrm{NAG}}\) is
\(\lambda^2-(a_{\mathrm{N}}+\bout)\lambda+D_{\mathrm{N}}\), so by
Cayley--Hamilton the same projection argument as in
\S\ref{app:restart-recurrence} gives
\[
    \chi_K^{\mathrm{NAG}}
    =
    (a_{\mathrm{N}}+\bout)\,\chi_{K-1}^{\mathrm{NAG}}
    -D_{\mathrm{N}}\,\chi_{K-2}^{\mathrm{NAG}},
    \qquad
    \chi_0^{\mathrm{NAG}}=1,\qquad
    \chi_1^{\mathrm{NAG}}=a_{\mathrm{N}}.
\]
When \(D_{\mathrm{N}}>0\) and
\((a_{\mathrm{N}}+\bout)^2<4D_{\mathrm{N}}\), the eigenvalues are complex
conjugates \(\rho_{\mathrm{N}}e^{\pm i\varphi_{\mathrm{N}}}\) with
\[
    \rho_{\mathrm{N}}=\sqrt{D_{\mathrm{N}}},
    \qquad
    \cos\varphi_{\mathrm{N}}
    =
    \frac{a_{\mathrm{N}}+\bout}{2\rho_{\mathrm{N}}}.
\]
Repeating the matching argument of \S\ref{app:closed-form} yields the same
structural form as for HB:
\[
    \chi_K^{\mathrm{NAG}}(\sigma)
    =
    \rho_{\mathrm{N}}^K\left[
    \cos(K\varphi_{\mathrm{N}})+
    C_{\mathrm{N}}\sin(K\varphi_{\mathrm{N}})
    \right],
    \qquad
    C_{\mathrm{N}}
    =
    \frac{a_{\mathrm{N}}-\bout}{2\rho_{\mathrm{N}}\sin\varphi_{\mathrm{N}}}.
\]
The structure is identical to the HB case, with three quantities replaced:
\(\bout\to D_{\mathrm{N}}\) in the determinant (and hence
\(\rho\to\rho_{\mathrm{N}}\)), \(a\to a_{\mathrm{N}}\) in the trace, and the
phase \(\varphi\) updated accordingly.

This makes the relationship between NAG and restarting explicit. NAG changes
the spectral envelope through $D_{\mathrm{N}}$, which now depends on
$\nu\sigma$; restarting instead acts on the phase-dependent projection factor
multiplying that envelope. The same projected-factor decomposition holds for
both HB and NAG. Thus, restarting can also be applied on top of NAG: when an
integer period $K$ places the NAG phase $K\varphi_{\mathrm{N}}$ near a
cancellation point, the NAG envelope $\rho_{\mathrm{N}}^K$ is multiplied by a
small oscillatory factor. In this sense, restarting can further contract the
NAG dynamics through the same cancellation mechanism as in HB, but with
NAG's phase $\varphi_{\mathrm{N}}$ instead of HB's phase $\varphi$. The two
mechanisms are therefore complementary: NAG improves the spectral envelope,
while restarting exploits favorable phases within that envelope.

\subsection{Blockwise extension}
\label{app:blockwise}

The scalar theory in the main text is mode-wise. A direct per-mode restart
period is useful analytically but unrealistic algorithmically. Here we describe
a more practical relaxation: one restart period per parameter block.

We connect this blockwise view to the linearized squared-loss setting through
the duality between the empirical NTK and the Gauss--Newton Hessian. If
\(J=[J_1,\ldots,J_B]\) is the Jacobian partitioned by parameter blocks, then
the empirical-kernel operator \(JJ^\top/D\) and the parameter-space
Gauss--Newton Hessian \(J^\top J/D\) have the same nonzero spectrum. Thus the
same effective-eigenvalue analysis can be read in parameter space, where
blockwise curvature structure is meaningful. Motivated by evidence that
Transformer Hessian spectra vary substantially across parameter blocks
\cite{zhang2024transformersadam}, and by work on near-block-diagonal Hessian
structure in neural networks \cite{dong2025hessianstructure}, assume
\begin{equation}
H_\theta^{\mathrm{GN}}
=
\frac1D J^\top J
\approx
\operatorname{blkdiag}(H_1,\ldots,H_B),
\qquad
H_b=\frac1D J_b^\top J_b .
\label{eq:block_gn_hessian}
\end{equation}
Equivalently, cross-block Gram terms \(J_b^\top J_c/D\) are small for
\(b\neq c\).

Under this approximation, block \(b\) evolves under its local curvature. After
\(S\) inner GD steps within outer round \(t\),
\begin{equation}
x_{t,b}^{\mathrm{loc}}
=
(I-\eta H_b)^Sx_{t,b},
\qquad
g_{t,b}
=
x_{t,b}-x_{t,b}^{\mathrm{loc}}
=
\Sigma_bx_{t,b},
\qquad
\Sigma_b:=I-(I-\eta H_b)^S .
\label{eq:block_pseudograd}
\end{equation}
Thus one outer transition already includes the \(S\) inner steps through
\(\Sigma_b\). The HB/EMA outer update in block \(b\) is therefore the same
two-state system as in the scalar case, with \(\sigma\) replaced by the block
operator \(\Sigma_b\):
\begin{equation}
\begin{pmatrix}
x_{t+1,b}\\
m_{t+1,b}
\end{pmatrix}
=
\begin{pmatrix}
I-\nu_b(1-\beta_b)\Sigma_b & -\nu_b\beta_b I\\
(1-\beta_b)\Sigma_b & \beta_b I
\end{pmatrix}
\begin{pmatrix}
x_{t,b}\\
m_{t,b}
\end{pmatrix}.
\label{eq:block_transition}
\end{equation}
Diagonalizing \(H_bu_{b,j}=\lambda_{b,j}u_{b,j}\) recovers independent scalar
dynamics inside the block, with
\begin{equation}
\sigma_{b,j}
=
1-(1-\eta\lambda_{b,j})^S .
\label{eq:block_effective_sigma}
\end{equation}
Thus a blockwise restart period \(K_b\) applies
\begin{equation}
x_{K_b,b}
=
\chi_{K_b}(\Sigma_b)x_{0,b}.
\label{eq:block_functional_contraction}
\end{equation}
Writing \(x_{0,b}=\sum_j \alpha_{b,j}u_{b,j}\), this becomes
\begin{equation}
x_{K_b,b}
=
\sum_j
\chi_{K_b}(\sigma_{b,j})\alpha_{b,j}u_{b,j}.
\label{eq:block_eigen_expansion}
\end{equation}
A natural blockwise oracle period is therefore
\begin{equation}
K_b^\star
\in
\argmin_K
\sum_j w_{b,j}|\chi_K(\sigma_{b,j})|^2,
\label{eq:blockwise_oracle}
\end{equation}
where \(w_{b,j}\) measures the residual energy in eigendirection \(j\) of block
\(b\). This gives a practical middle ground between one global period and an
unrealistic per-eigenmode oracle.

\paragraph{High-momentum period heuristic.}
If the effective spectrum of block \(b\) is concentrated near
\(\bar\sigma_b\), and \(\beta_b=1-\delta_b\) with \(\delta_b\ll1\), the scalar
phase satisfies
\begin{equation}
\varphi_b
\approx
\sqrt{\nu_b\bar\sigma_b(1-\beta_b)} .
\label{eq:block_phase_expansion}
\end{equation}
Indeed, using
\(a=1-\nu_b(1-\beta_b)\bar\sigma_b\) and
\(\rho=\sqrt{\beta_b}\), we have
\[
\cos\varphi_b
=
\frac{a+\beta_b}{2\rho}
=
1-\frac{\nu_b\bar\sigma_b(1-\beta_b)}{2}
+O\bigl((1-\beta_b)^2\bigr),
\]
which gives \eqref{eq:block_phase_expansion}. The phase shift
\(\theta=\arctan C\) is \(O(\sqrt{1-\beta_b})\), so the first cancellation
period satisfies
\begin{equation}
K_b
\approx
\frac{\pi}{2\varphi_b}
\approx
\frac{\pi}{2\sqrt{\nu_b\bar\sigma_b(1-\beta_b)}} .
\label{eq:block_period_heuristic}
\end{equation}
Thus blocks with larger effective progress \(\bar\sigma_b\) prefer shorter
restart periods, while slower blocks prefer longer ones.
\clearpage
\newpage

\section{Experimental Details and Additional Results}
\label{app:sec:experiments}

In this appendix, we detail the additional experiments completed as well as present more detailed versions of plots from the main body. In addition to this, we show hyperparameter configurations used.

\subsection{Details for Figure~\ref{fig:restart_mechanism}.}
All panels use \(\bout=0.9\) and \(\nu=1\), except panel (a), which also shows a
no-restart NAG baseline with \(\bout\) tuned to minimize the final loss at the
same horizon. Panel (a) uses a single mode with \(\sigma=0.95\), horizon
\(T=80\), and initial state \(z_0=(x_0,m_0)=(1,0)\). Panel (b) uses the
multi-mode spectrum
\(\sigma\in\{0.95,0.85,0.75,0.60,0.45,0.30\}\), equal weights, and horizon
\(T=120\); the per-mode restart curve is an oracle benchmark. Panel (c) uses
three heterogeneous blocks with effective spectra concentrated in
\([0.92,1.00]\), \([0.55,0.65]\), and \([0.18,0.26]\), again with equal
weighting and horizon \(T=120\).

\begin{figure*}[htp]
\centering
\includegraphics[width=0.95\textwidth]{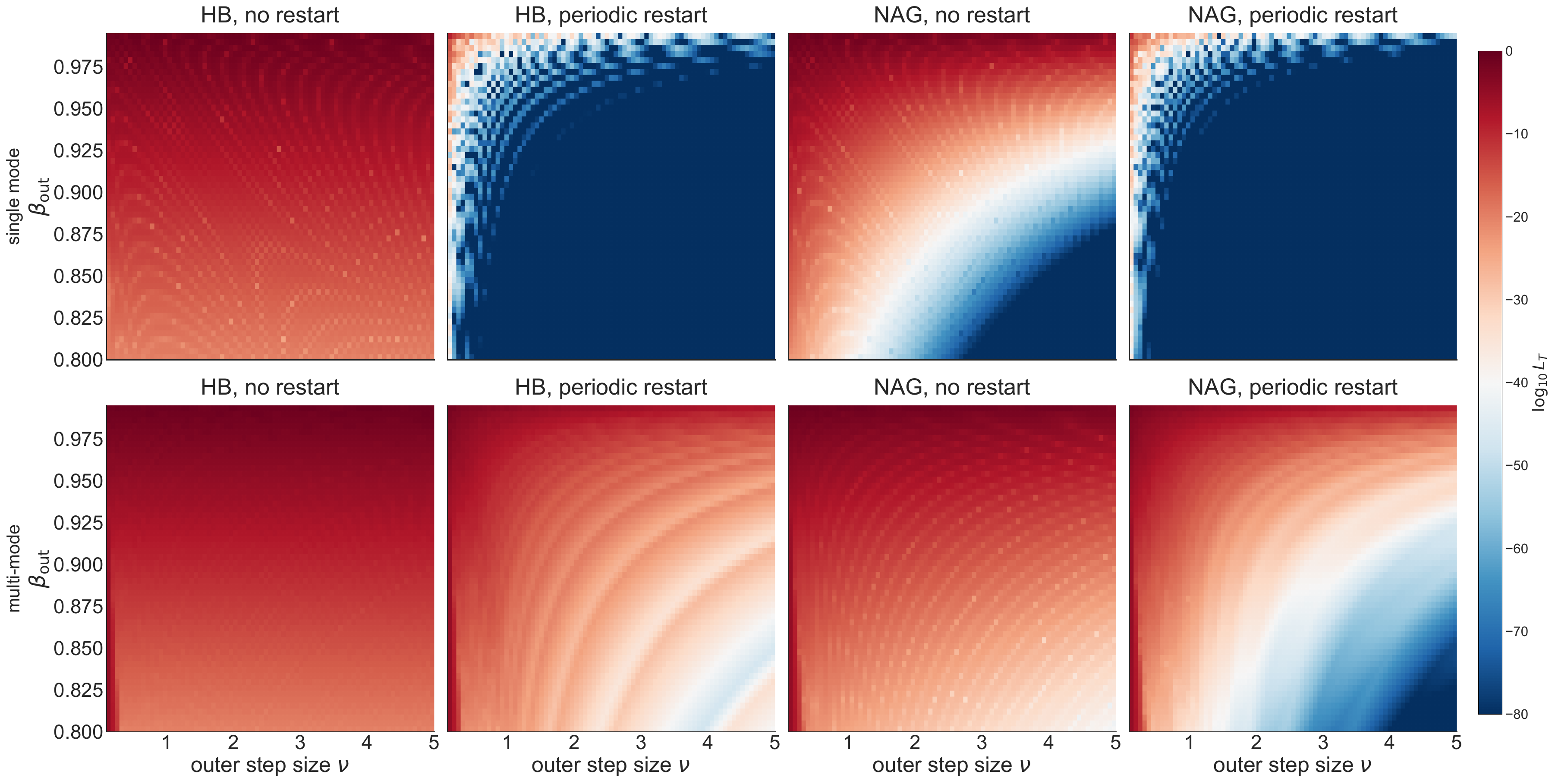}
\vspace{-2mm}
\caption{\textbf{Robustness over outer hyperparameters.}
Each heatmap shows clipped \(\log_{10}\) final loss over a grid of \(\bout\)
and \(\nu\). The best periodic restart enlarges the good-hyperparameter region for HB and NAG.}
\label{fig:robustness}
\vspace{-2mm}
\end{figure*}

\subsection{Hyperparameter Sweeps for Llama-150M}

We evaluate DiLoCo by pretraining Llama-150M on the C4 dataset from scratch using a 2-replica DiLoCo configuration with two H200 GPUs (128GB memory). We first tuned the inner optimizer's learning rate and carried the optimal learning rate of 
$1 \times 10^{-3}$ 
forward to the rest of the experiments. A full grid search over all combinations of hyperparameters would be computationally prohibitive, so experiments were designed to isolate one or two axes of variation at a time while holding the others fixed.

\label{app:experimental_details}

\begin{table}[htbp]
\centering
\small
\setlength{\tabcolsep}{5pt}
\renewcommand{\arraystretch}{1.12}
\caption{Hyperparameter configuration for the DiLoCo runs.}
\label{tab:diloco_experimental_setup}
\begin{tabular}{llp{0.43\linewidth}}
\toprule
\textbf{Category} & \textbf{Hyperparameter} & \textbf{Value} \\
\midrule
\multirow{3}{*}{Batching}
  & Per-replica effective batch size & 64 \\
  & Effective global batch size      & 128 \\
\midrule
\multirow{2}{*}{Inner optimizer}
  & Learning rate (AdamW)            & $5{\times}10^{-4},\;1{\times}10^{-3},\;2{\times}10^{-3}$ \\
  & Communication period $S$         & 64, 128, 512, 1024, 2048 \\
\midrule
\multirow{4}{*}{Outer optimizer}
  & Type                             & Heavy-ball, Nesterov momentum \\
  & Learning rate $\nu$              & 0.1, 0.3, 0.5, 0.7, 0.9, 1.1 \\
  & Momentum $\beta_{\mathrm{out}}$  & 0.1, 0.3, 0.5, 0.7, 0.9 \\
  & Restart period $K$               & 2, 3, 5, 7, 9, 11 \\
\midrule
\multirow{3}{*}{Training schedule}
  & LR scheduler                     & Cosine annealing \\
  & Total inner steps                & 12{,}500 \\
  & Warmup steps                     & 1{,}200 \\
\bottomrule
\end{tabular}
\end{table}

In all experiments, the inner optimizer is AdamW with weight decay $0.1$ and betas $(0.9,0.999)$.

\begin{table}[t]
\centering
\small
\setlength{\tabcolsep}{5pt}
\renewcommand{\arraystretch}{1.12}
\caption{Model, data, and evaluation configuration for DiLoCo experiments.}
\label{tab:model_data_eval_config}
\begin{tabular}{llp{0.43\linewidth}}
\toprule
\textbf{Category} & \textbf{Configuration} & \textbf{Value} \\
\midrule
\multirow{5}{*}{Model}
  & Architecture          & \texttt{LlamaForCausalLM}$^{\dagger}$ \\
  & Hidden size           & 1024 \\
  & Intermediate size     & 2688 \\
  & Hidden layers         & 12 \\
  & Attention heads       & 16 \\
  & RMSNorm $\epsilon$    & $10^{-5}$ \\
\midrule
\multirow{3}{*}{Data}
  & Dataset               & \texttt{allenai/c4} \\
  & Tokenizer             & \texttt{Mistral-7B-v0.1}$^{\ddagger}$ \\
  & Sequence length       & 2048 \\
\midrule
\multirow{3}{*}{Runtime}
  & Precision             & \texttt{bf16-mixed} \\
  & Compute               & 2 replicas, 2 H200 GPUs \\
\bottomrule
\end{tabular}
\\[4pt]
\raggedright\footnotesize
$^{\dagger}$\texttt{PrimeIntellect/llama-150m-fresh}\quad
$^{\ddagger}$\texttt{mistralai/Mistral-7B-v0.1}
\end{table}

\subsection{Soft Restarts}

\begin{figure}[H]
    \centering
    \begin{minipage}{0.48\textwidth}
        \centering
        \includegraphics[width=\linewidth]{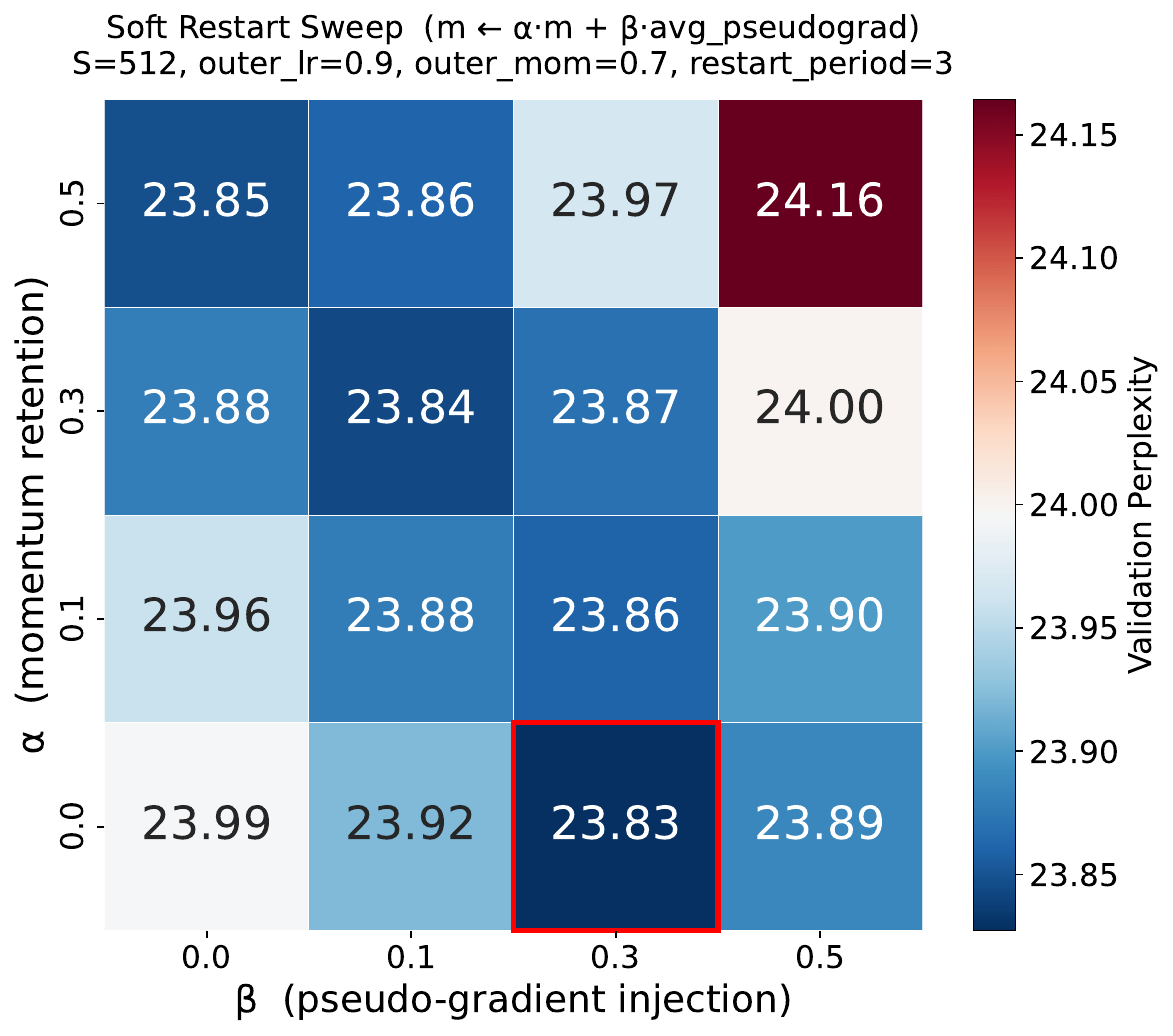}
        \caption*{\textbf{(a)} \(S=512\), \(\nu=0.9\), \(\beta_{\text{out}}=0.7\), \(K=3\).}
    \end{minipage}
    \hfill
    \begin{minipage}{0.48\textwidth}
        \centering
        \includegraphics[width=\linewidth]{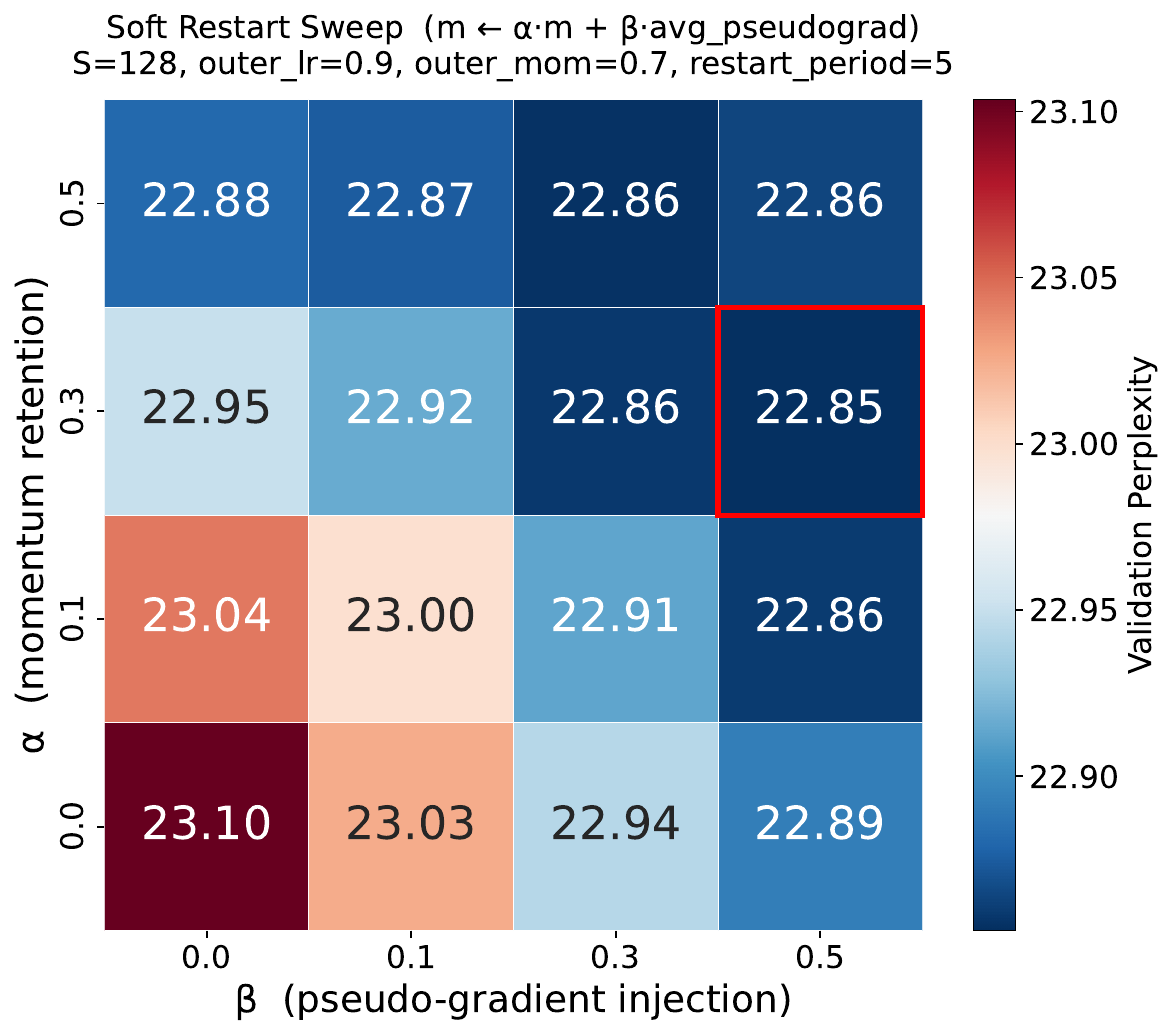}
        \caption*{\textbf{(b)} \(S=128\), \(\nu=0.9\), \(\beta_{\text{out}}=0.7\), \(K=5\).}
    \end{minipage}
    \caption{Soft restart sweeps. Each cell shows final validation perplexity under the boundary update
    \(m \leftarrow \alpha m + \beta \bar g\), where \(\alpha\) controls momentum retention and
    \(\beta\) controls pseudo-gradient injection at the restart boundary.}
    \label{fig:soft_restart_heatmaps}
\end{figure}

A soft restart applies a periodic buffer rewrite on top of the ordinary outer momentum update, occurring only at restart boundaries.
Unlike a hard restart, which completely discards the accumulated momentum state by zeroing the buffer, the soft restart can be viewed as partially preserving the existing momentum while incorporating the current averaged DiLoCo pseudogradient.
Specifically, at a restart boundary, we rewrite the buffer as
$ \widetilde{m}_t \leftarrow \alpha m_t + \beta \bar{g_t}$,
where $m_t$ denotes the momentum buffer after the ordinary outer optimizer update but before any restart rewrite, $\widetilde{m}_t$ denotes the momentum buffer after the possible soft-restart rewrite, and $\bar{g_t}$ denotes the averaged DiLoCo pseudogradient.
Here, $\alpha$ governs how much of existing momentum memory we want to retain, while $\beta$ governs the contribution of the current averaged pseudogradient to the rewritten buffer. We define it as
$$
\widetilde{m}_t =
\begin{cases}
\alpha m_t + \beta \bar{g}_t, & t \equiv 0 \pmod{R} \\
m_t, & \text{o.w.}
\end{cases} \qquad \text{for } t \in \{1,\ldots,T\}.
$$
where $R$ denotes the restart period in outer steps. Below we present Figure~\ref{fig:soft_restart_heatmaps} that illustrates how the optimal $(\alpha, \beta)$ landscape shifts substantially with the training configuration between $S=512$ and $S=128$. While carefully tuned soft restarts can marginally improve over hard restarts and sometimes improve over the optimal no-restart, the improvements are small in absolute perplexity terms and come at the cost of two additional hyperparameters whose optimal values appear to be sensitive to $S$ and $K$. Given that the simpler hard restart already provides the robustness benefits described in Section~\ref{sec:experiments}, we do not find soft restarts to be worth the added tuning burden in practice.

\subsection{Additional Robustness Results for DiLoCo}
\label{app:robustness_curves}

\begin{figure}[H]
    \centering
    \includegraphics[width=1\textwidth]{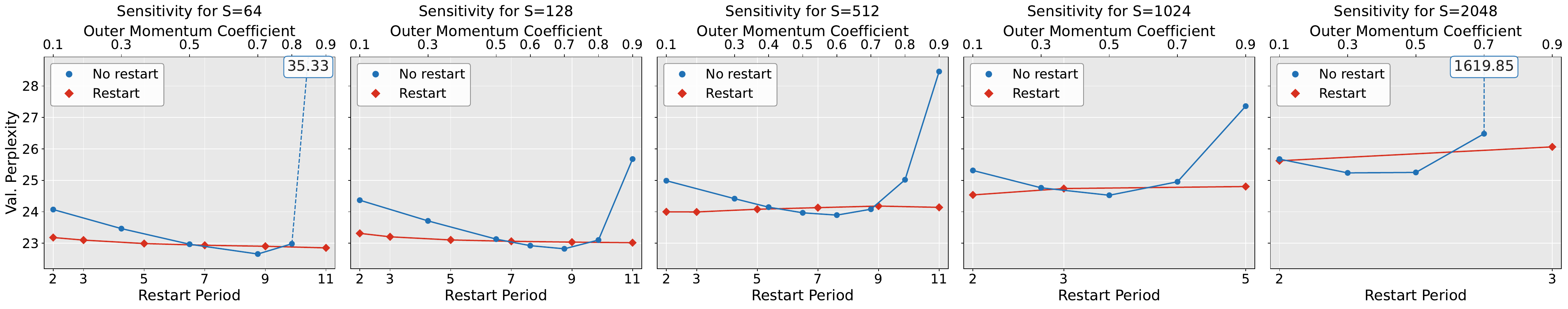}
    \caption{Final validation perplexity for \textbf{NAG} as outer optimizer, shown as a function of restart period $K$ (bottom axis, red) and outer momentum coefficient $\beta_{\text{out}}$ (top axis, blue) for $S \in \{64, 128, 512, 1024, 2048\}$. The blue curve sweeps $\beta_{\text{out}}$ with no momentum restarts; the red curve fixes $\beta_{\text{out}}$ and sweeps $K$. Dashed lines with annotated values indicate runs that diverged and exceeded the plot range. Without restarts, perplexity spikes sharply at high $\beta_{\text{out}}$, with catastrophic divergence at $S = 2048$; with restarts, the red curve remains nearly flat across all tested $K$, demonstrating that the restart period is a more forgiving hyperparameter than outer momentum.}
    \label{fig:robust_diloco_full}
\end{figure}

We use the term \textit{robustness} 
to mean low sensitivity, or equivalently low variation, 
of the validation metric 
across the explored hyperparameter settings.
Intuitively, robustness across hyperparameter choices is desirable because it reduces reliance on exhaustive tuning over a finely discretized hyperparameter grid to achieve good validation performance. Figure~\ref{fig:robust_diloco_full} extends the robustness comparison from Figure~\ref{fig:robust_diloco} to a wider range of communication periods, $S \in \{64, 128, 512, 1024, 2048\}$, reporting the final perplexity. Across all the communication frequencies, we see that the no-restart curve's optimal $\beta_{\text{out}}$ shift downwards as $S$ increases, as well as degrade sharply at high $\beta_{\text{out}}$, with divergence at $S=2048$. On the other hand, the restart curve remains nearly flat for all tested periods $K$.

\begin{figure}[H]
    \centering
    \includegraphics[width=1\textwidth]{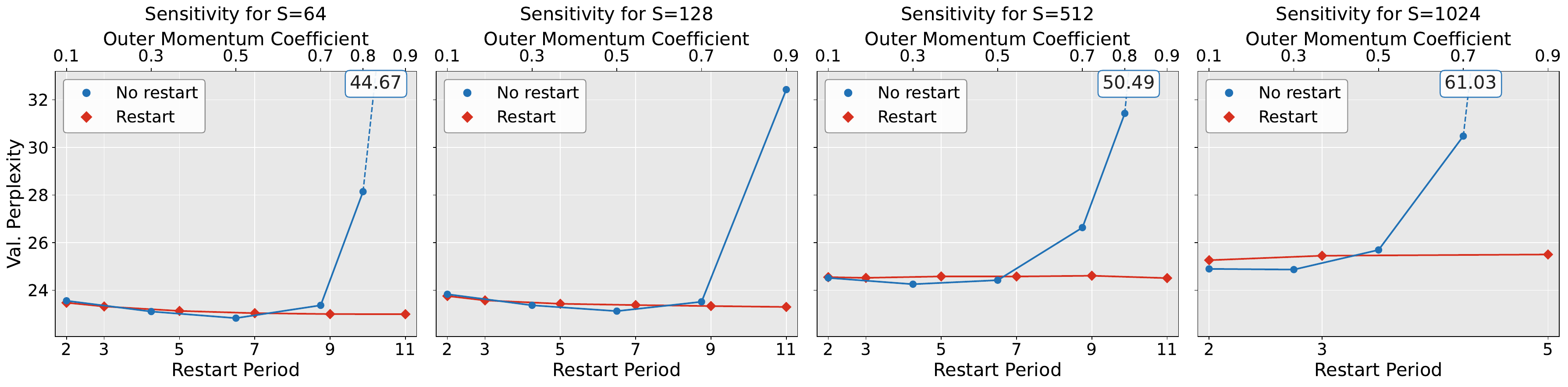}
    \caption{Same as Figure~\ref{fig:robust_diloco_full} but with \textbf{Heavy-Ball} (EMA) as the outer optimizer, for $S \in \{64, 128, 512, 1024\}$. Heavy-Ball without restarts is more sensitive to $\beta_{\text{out}}$ than NAG, with earlier and more severe divergence as $S$ increases. Periodic restarts again stabilize training across all communication periods, keeping perplexity flat across the sweep of $K$.}
    \label{fig:robust_heavyball_1}
\end{figure}

\begin{figure}[htp!]
    \centering
    \begin{minipage}{0.49\textwidth}
        \centering
        \includegraphics[width=\linewidth]{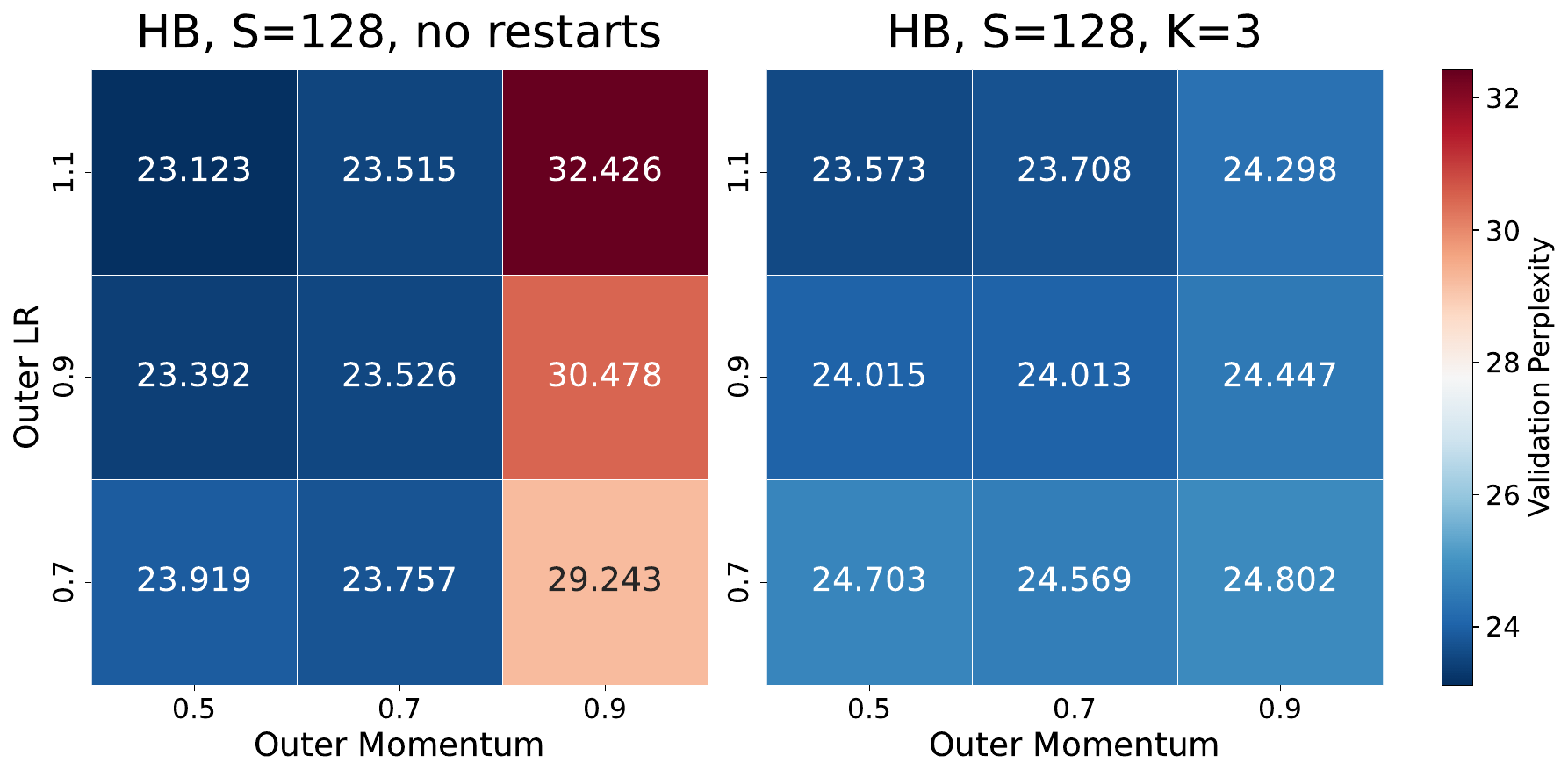}
    \end{minipage}
    \hfill
    \begin{minipage}{0.49\textwidth}
        \centering
        \includegraphics[width=\linewidth]{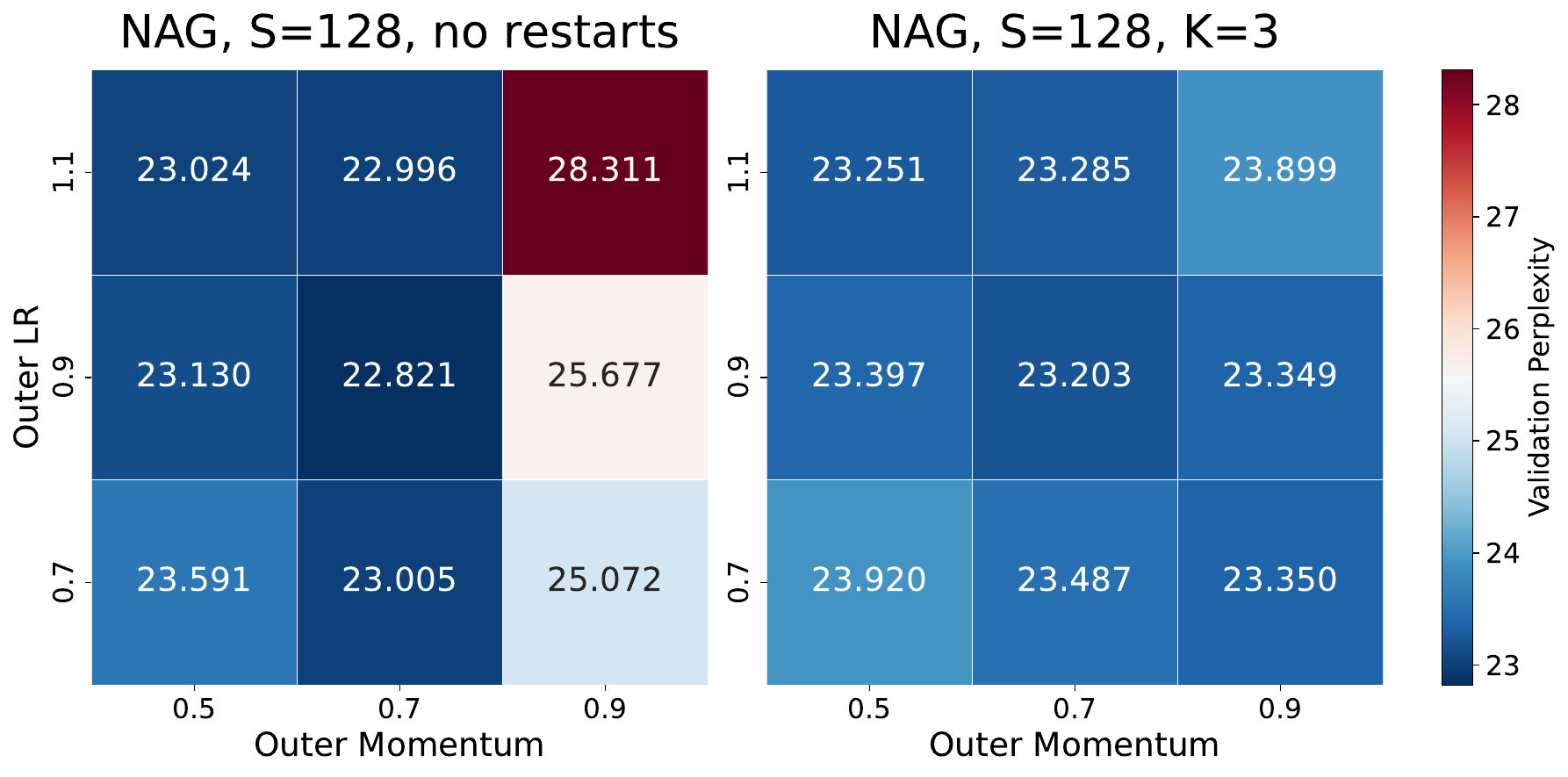}
    \end{minipage}
    \caption{Validation perplexity over the outer hyperparameter grid at \(S=128\) for HB(left) and NAG(right), comparing standard DiLoCo against DiLoCo with momentum restart period \(K=3\). Periodic restarts reduce the high-\(\beta_{\mathrm{out}}\) failure region while preserving peak performance.}
    \vspace{-1em}
    \label{fig:robust_heavyball_heatmap}
\end{figure}

\end{document}